\documentclass[10pt,twocolumn,letterpaper]{article}

\usepackage{cvpr}
\usepackage{times}
\usepackage{epsfig}
\usepackage{graphicx}
\usepackage{amsmath}
\usepackage{amssymb}

\usepackage{multirow}
\usepackage{amsfonts}
\usepackage{booktabs}
\usepackage{verbatim}
\usepackage{enumitem}
\usepackage{subfigure}
\usepackage{amsmath}
\usepackage{xcolor}
\usepackage{blindtext}
\usepackage{overpic}
\usepackage{transparent}

\usepackage[pagebackref=true,breaklinks=true,letterpaper=true,colorlinks,bookmarks=false]{hyperref}

\newcommand{\myparagraph}[1]{\medbreak\noindent\textbf{#1}}
\cvprfinalcopy

\def\cvprPaperID{712}

\ifcvprfinal\fi
\begin{document}

\title{FReeNet: Multi-Identity Face Reenactment}
\author{Jiangning Zhang$^1$\thanks{Work mainly done during an internship at NetEase Fuxi AI Lab.}~~\thanks{Equal contribution.}
~ ~ Xianfang Zeng$^1$\footnotemark[2]
~ ~ Mengmeng Wang$^1$\footnotemark[2]
~ ~ Yusu Pan$^1$ \\
~ ~ Liang Liu$^1$
~ ~ ~ ~ ~ ~ Yong Liu$^1$\thanks{Corresponding author.}
~ ~ ~ ~ ~ ~ ~ Yu Ding$^2$
~ ~ ~ ~ ~ ~ Changjie Fan$^2$ \\
\normalsize $^1$ Zhejiang University ~ ~ $^2$Fuxi AI Lab, NetEase \\
{\tt\small \{186368, zzlongjuanfeng, mengmengwang, corenel, leonliuz\}@zju.edu.cn} \\
{\tt\small yongliu@iipc.zju.edu.cn, \{dingyu01, fanchangjie\}@corp.netease.com}
}
\maketitle
\thispagestyle{empty}

\begin{abstract}
    This paper presents a novel multi-identity face reenactment framework, named FReeNet, to transfer facial expressions from an arbitrary source face to a target face with a shared model.
    The proposed FReeNet consists of two parts: Unified Landmark Converter (ULC) and Geometry-aware Generator (GAG).
    The ULC adopts an encode-decoder architecture to efficiently convert expression in a latent landmark space, which significantly narrows the gap of the face contour between source and target identities. 
    The GAG leverages the converted landmark to reenact the photorealistic image with a reference image of the target person.
    Moreover, a new triplet perceptual loss is proposed to force the GAG module to learn appearance and geometry information simultaneously, which also enriches facial details of the reenacted images.
    Further experiments demonstrate the superiority of our approach for generating photorealistic and expression-alike faces, as well as the flexibility for transferring facial expressions between identities.
\vspace{-1em}
\end{abstract}

\section{Introduction}
Face reenactment is a task to transfer the facial expression from one source face to a target face, which has vast promising applications such as film-making, facial animations, and augmented reality.
Moreover, a live video of a particular person can be generated under the control of another person, which can attack or strengthen the bioassay system.
In this work, we focus on solving a more challenging task: multi-identity face reenactment, where the source face is from an arbitrary person and the target person is not specific.
This task is distinguished from one(many)-to-one face reenactment tasks in whether the target is specific, which is more general and flexible in practical applications.

Benefiting from the release of the large-scale face datasets~\cite{LFWTech,LFWTechUpdate,gross2010multi,yang2016wider}, many high-accuracy and reliable face detection algorithms are proposed~\cite{wu2018look,lv2017deep,dong2018style,guo2019pfld} that contribute to the development of the face reenactment task.
Many excellent facial expression migration and face reenactment methods have been proposed in the last decade, which summarily fall into two main categories: three dimensional (3D) model based synthesis and GAN based generation.
For the 3D model based synthesis methods~\cite{vlasic2005face,kim2018deep,garrido2016reconstruction,suwajanakorn2017synthesizing,thies2018headon}, the person is represented by a predefined parametric model.
Generally, the methods firstly capture the facial movement of a source video that will be fitted into a parametric space over the predefined model and then render the target video through morphing.
These techniques are famous for the animation of computer graphics (CG) avatars in both games and movies~\cite{pumarola2018ganimation} because they have high-quality and high-resolution face reenactment ability.
However, these methods generally suffer from big-budget model making and are computationally expensive.

\begin{figure}[t]
    \centering
    \includegraphics[width=1.0\columnwidth]{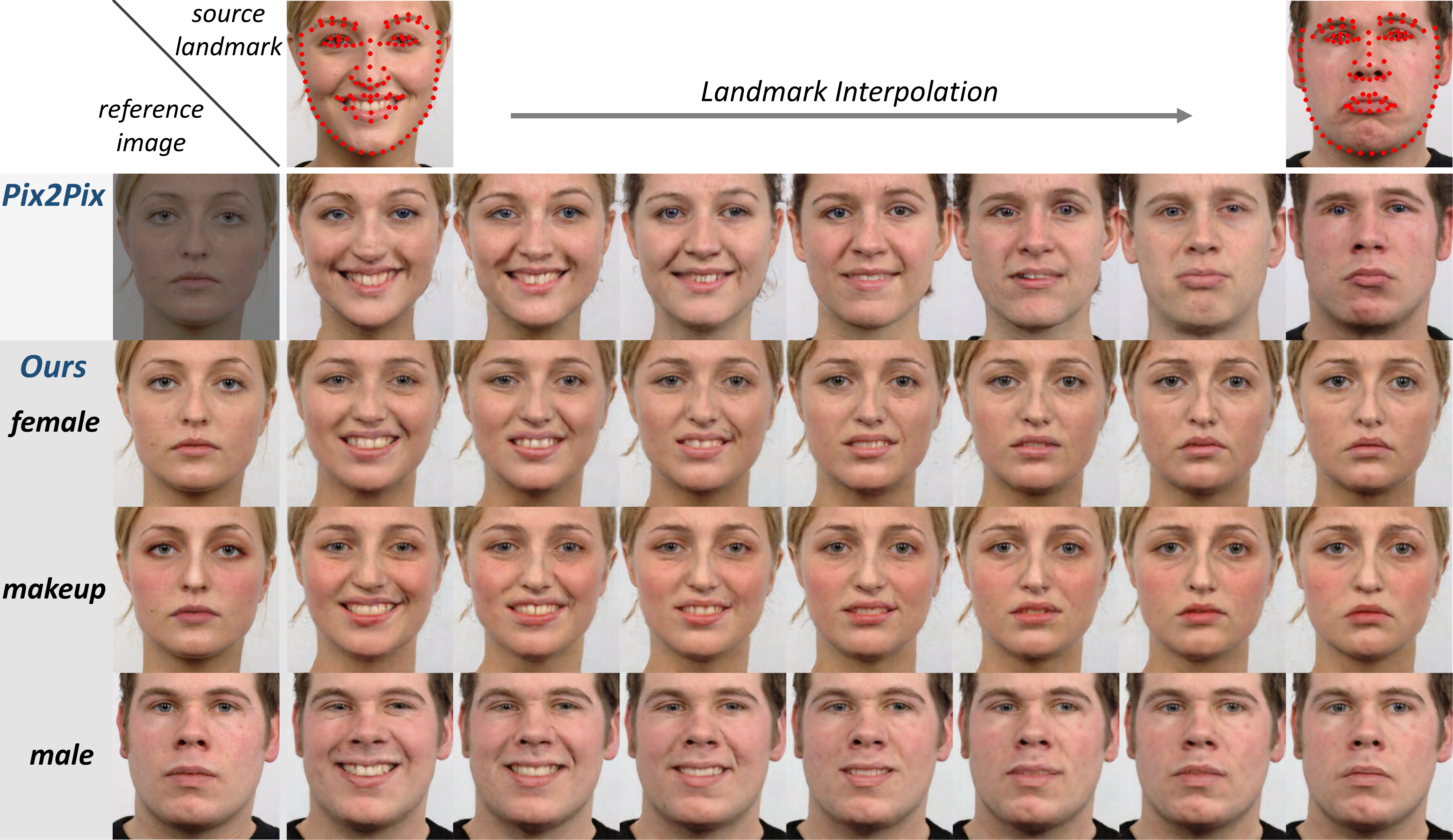}
    \caption{A toy experiment of the face reenactment task. The results of the first row are from \emph{Pix2Pix}~\cite{pix2pix2017} method with the only landmark as input, while the second to fourth rows are from our method with different reference faces but the same geometry information. Note that the reference image of the third row is the makeup of the girl in the second row.
    }
    \vspace{-1.5em}
    \label{fig:1-reference_comparison}
\end{figure}

Recently, many GAN based methods have achieved a significant improvement~\cite{geng2018warp,jin2017cyclegan,xu2017face,wu2018reenactgan,pumarola2018ganimation,wiles2018x2face} due to the natural advantages in learning distribution patterns from a large-scale dataset.
Generally, the encoder-decoder structure is adopted to reenact target face in an adversarial idea~\cite{pix2pix2017}, and some further works~\cite{CycleGAN2017,jin2017cyclegan,song2018geometry,xu2017face} introduce a powerful cycle consistent loss to achieve unpaired face reenactment.
However, they can only reenact faces between two specific identities once the network is trained.
Subsequently, the many-to-one face reenactment task sets out to solve the above problem, which can reenact one target face from multiple persons using only one network.
Recent Nirkin~\etal~\cite{Fsgan} derive a recurrent neural network based approach for the subject agnostic face reenactment, but it requires extra segmentation information and has much more parameters than common methods.
ReenactGAN~\cite{wu2018reenactgan} introduces a \emph{Transformer} module to adapt source facial movements from multiple persons to the target person in a latent facial boundary space, and then decodes the target face.
Nonetheless, it is still network-inefficient in practical applications for requiring a transformer and a decoder for each target person. 
So it is of considerable significance to implement a multi-identity face reenactment task, also called many-to-many face reenactment, in a unified network, where both the source and reference faces can be from multiple persons.
X2Face~\cite{xu2017face} achieves the task by first adopting an embedding network to encode an embedded face and then using a driving network to reenact the target face, but the generative images are not satisfying in quality and facial details. 
In summary, there are still two main challenges in this task:
(1) How to convert multi-identity facial expression by a unified network, because there exists a gap in facial contours between source and target persons.
(2) How to reenact photorealistic and identity-consistent target face as the reference face while keeping consistent with the pose, hue, and illumination.  

To address the issues above, we propose a multi-identity face reenactment framework named FReeNet, which can efficiently transfer expressions from arbitrary source identities to target identities.
Firstly, a landmark detector~\cite{guo2019pfld} is leveraged to encode faces into a latent landmark space.
This latent space serves as a high-quality bridge for the next facial converting step, where the facial geometry information is efficiently preserved while the appearance information is omitted.
Subsequently, a unified landmark converter module is applied to effectively convert the expression of an arbitrary source person to the target person in latent landmark space. Then the geometry-aware generator simultaneously extracts geometry information from the converted landmark and appearance information from the reference person to reenact the target face.
Such a decoupling design can generate distinguishing faces with similar or same landmark inputs, because different persons with different appearances are used as the reference face.
Furthermore, we combine the triplet loss~\cite{schroff2015facenet} and the perceptual loss~\cite{johnson2016perceptual} to form a novel triplet perceptual loss, which helps generate more detailed images as well as decouple the appearance and geometry information.
As shown in Figure~\ref{fig:1-reference_comparison}, our approach can effectively preserve the reference identity while converting the geometry information.

To our best knowledge, the proposed FReeNet is the first to successfully perform multi-identity face reenactment task using a unified model, while keeping the pose, hue, and illumination information consistent with the reference face.
Specifically, we make the following four contributions:
\vspace{-1mm}
\begin{itemize}
\vspace{-1mm}
\item A unified landmark converter is proposed to convert the expression from the source identity to the target identity, and both the source and target identities are from multiple persons.
\vspace{-1mm}
\item A geometry-aware generator is proposed to reenact the photorealistic target face, which is designed in a decoupling idea and extracts the appearance and geometry information from separate paths.
\vspace{-1mm}
\item A new triplet perceptual loss is proposed to enrich the facial details of the reenacted face.
\vspace{-1mm}
\item Experimental results indicate that the proposed approach implements the many-to-many face reenactment task and can generate high-quality and identity-consistent face images.
\end{itemize}
\vspace{-2mm}

\section{Related Work}
\myparagraph{Image Synthesis.}
Driven by remarkable generation effects of GAN~\cite{goodfellow2014generative}, researchers have achieved excellent results in various domains, such as
image translation~\cite{radford2015unsupervised,pix2pix2017,CycleGAN2017,wang2018pix2pixHD},
person image synthesis~\cite{dong2018soft,ma2017pose}, and
face generation~\cite{StarGAN2018,pumarola2018ganimation,karras2017progressive,karras2018style}.
Mehdi~\etal~\cite{mirza2014conditional} designed a cGAN structure to condition on to both the generator and discriminator for a more controllable generation of attributes.
Subsequently, the Pix2Pix~\cite{pix2pix2017} achieved incredible results in paired image translation tasks by using L1 and adversarial losses between the generated image and the ground-truth.
Zhu~\etal~\cite{CycleGAN2017} subsequently proposed a new cycle consistency loss for unpaired image translation between two domains, which dramatically reduces the requirement of the data annotation.
DualGAN~\cite{yi2017dualgan} analogously learns two translators from one domain to the other and hence can solve general-purpose image-to-image translation tasks.
Furthermore, StarGAN~\cite{StarGAN2018} proposed a unified model for multi-domain facial attribute transferring and expression synthesis.
Recently, some methods can generate vivid faces directly from the latent input code. Tero~\etal~\cite{karras2017progressive} described a new progressive growing training methodology for the face generation from an underlying code.
StyleGAN~\cite{karras2018style} proposed a style-based generator that embeds the latent input code into an intermediate latent space, which can control the weights of image features at different scales and synthesize extremely naturalistic face images.
However, using latent code as the input is an uncontrollable generation process that is not suitable for the face synthesis task, and they have limited scalability in handling the many-to-many face reenactment task.
Our method introduces a landmark space for expression transferring among multiple persons and uses converted landmark images as the guidance to reenact target faces, which is not the same as the existing methods.

\myparagraph{Face Reenactment.}
Benefiting from large-scale face database collections~\cite{LFWTech,LFWTechUpdate,yang2016wider} and reliable landmark detectors~\cite{guo2019pfld,wu2018look,dong2018style,lv2017deep}, numerous impressive face reenactment methods have been proposed in recent years.
These methods can be roughly categorized into either 3D model based method or GAN based method.
In the 3D model based branch, Volker~\etal~\cite{blanz1999morphable} proposed the morphable 3D model to estimate the parameters of the shape-vector and texture-vector, which are then used to recover full 3D shape and texture of the face.
Subsequent Face2face~\cite{thies2016face2face} applied an efficient deformation transfer to tracked facial expressions of both source and target videos, and then re-rendered the synthesized target faces for better fitting the retrieved and warped mouth interior.
Ma~\etal~\cite{ma2019real} reconstructed high-resolution facial geometry and appearance by capturing an individual-specific face model with fine-scale details.
Those 3D model based methods generally require delicately designed models, which are time and money consuming, and they are also computationally expensive. 
Therefore, Albert~\etal~\cite{pumarola2018ganimation} introduced GANimation to control the magnitude of activation of each AU and then combine several of them to synthesize target faces, which only requires images without other procedures.
Jin~\etal~\cite{jin2017cyclegan} directly applied CycleGAN to transfer facial expressions between two identities.
Recently, Wu~\etal proposed ReenactGAN~\cite{wu2018reenactgan} that is capable of transferring facial movements from one monocular image of multiple persons to a specific target person.
However, our framework aims at solving the harder many-to-many face reenactment problem using a unified concise framework, which has more promising applications.

\begin{figure*}[htb] 
    \centering
    \includegraphics[width=0.95\linewidth]{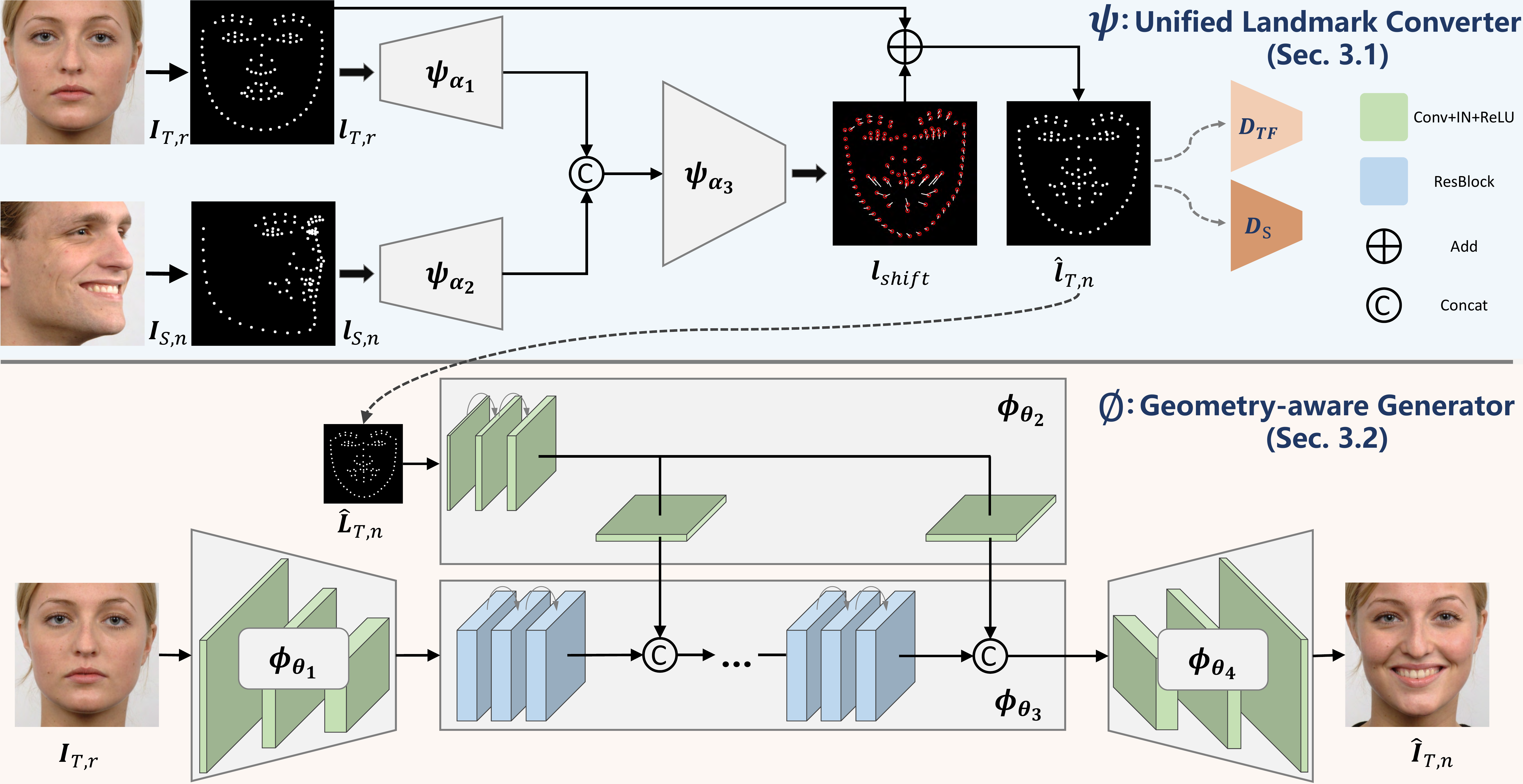}
    \caption{Overview of the proposed FReeNet. The net consists of a unified landmark converter $\boldsymbol{\psi}$ and a geometry-aware generator $\boldsymbol{\phi}$. Given a source person with an arbitrary expression $\boldsymbol{I}_{S,n}$ and a target person with the reference expression $\boldsymbol{I}_{T,r}$, the converter $\boldsymbol{\psi}$ uses extracted landmarks $\boldsymbol{l}_{S,n}$ and $\boldsymbol{l}_{T,r}$ to regress a landmark shift $\boldsymbol{l}_{shift}$, and then constructs the converted landmark $\hat{\boldsymbol{l}}_{T,n}$. Two discriminators $\boldsymbol{D}_{TF}$ and $\boldsymbol{D}_{S}$ are adopted for adversarial training. After that, geometry-aware generator $\boldsymbol{\phi}$ reenacts target face $\hat{\boldsymbol{I}}_{T,n}$ under the guidance of $\hat{\boldsymbol{L}}_{T,n}$, where $\boldsymbol{I}_{T,r}$ is used as the reference image. $\alpha_{i}(i=1,2,3)$ and $\theta_{j}(j=1,2,3,4)$ represent partial parameters of the network.}
    \vspace{-1em}
    \label{fig:3-FReeNet}
\end{figure*}

\section{FReeNet}
In this paper, a novel framework named FReeNet is proposed to complete the multi-identity face reenactment task efficiently.
As depicted in Figure~\ref{fig:3-FReeNet}, we first adopt the face landmark detector~\cite{guo2019pfld} to encode two input images $\boldsymbol{I}_{T,r}$ and $\boldsymbol{I}_{S,n}$ ($\in\mathbb{R}^{3\times 256\times 256}$) to a latent landmark space $\boldsymbol{l}_{T,r}$ and $\boldsymbol{l}_{S,n}$ ($\in\mathbb{R}^{106\times 2}$), where the first subscript means identity ($T$ means target person and $S$ means source person) and the second is expression ($r$ means the reference expression and $n$ means an arbitrary expression).
For example, the $\boldsymbol{l}_{T,r}$ represents the landmark from the target person with the reference expression (the neutral expression is used as the reference expression in the paper) that can be in a different pose.
The unified landmark converter subsequently adapts the source expression to the target, denoted as $\boldsymbol{\psi}:(\boldsymbol{l}_{T,r}, \boldsymbol{l}_{S,n}) \rightarrow \hat{\boldsymbol{l}}_{T,n}$.
Finally, the geometry-aware generator simultaneously leverages the converted geometry information $\hat{\boldsymbol{L}}_{T,n}\in\mathbb{R}^{1\times 64\times 64}$ and the appearance information $\boldsymbol{I}_{T,r}\in\mathbb{R}^{3\times 256\times 256}$ to reenact the target face $\hat{\boldsymbol{I}}_{T,n}\in\mathbb{R}^{3\times 256\times 256}$, denoted as $\boldsymbol{\boldsymbol{\phi}}: (\hat{\boldsymbol{L}}_{T,n}, \boldsymbol{I}_{T,r}) \rightarrow \hat{\boldsymbol{I}}_{T,n}$.
$\hat{\boldsymbol{L}}_{T,n}$ represents plotted landmark image from generated landmark vector $\hat{\boldsymbol{l}}_{T,n}$.
Moreover, a new triplet perceptual loss is introduced to boost the performance of the GAG.

\subsection{Unified Landmark Converter}
As mentioned in~\cite{wu2018reenactgan}, it may lead to artifacts if we directly apply ill-suited facial contour to synthesis the target image.
In contrast to existing methods, we design a unified landmark converter module (ULC) to adapt the source expression from an arbitrary person to the target person.
It can significantly alleviate the geometrical gap between the source and target faces.
As shown in Figure~\ref{fig:3-FReeNet} (top), the proposed ULC module contains two landmark encoders ($\boldsymbol{\psi}_{\alpha_1}$ and $\boldsymbol{\psi}_{\alpha_2}$) and a landmark shift decoder ($\boldsymbol{\psi}_{\alpha_3}$). 
Encoders $\boldsymbol{\psi}_{\alpha_1}$ and $\boldsymbol{\psi}_{\alpha_2}$ extract landmark features of the target and source faces respectively, and then the decoder $\boldsymbol{\psi}_{\alpha_3}$ fuses them to estimate the landmark shift $\boldsymbol{l}_{shift}$.
After that, we add $\boldsymbol{l}_{T,r}$ and $\boldsymbol{l}_{shift}$ in a point-wise manner to get the converted landmark $\hat{\boldsymbol{l}}_{T,n}$, which has the facial contour with $\boldsymbol{I}_{T,r}$ while keeping the expression information of $\boldsymbol{I}_{S,n}$.
This process is denoted as:

\vspace{-.5em}
\begin{equation}
  \begin{aligned}
    \hat{l}_{T,n} = & \boldsymbol{l}_{T,r} + \boldsymbol{l}_{shift}  \\
    = & \boldsymbol{l}_{T,r} + \boldsymbol{\psi}_{\alpha_3}(\boldsymbol{\psi}_{\alpha_1}(\boldsymbol{l}_{T,r}), \boldsymbol{\psi}_{\alpha_2}(\boldsymbol{l}_{S,n}))  \text{.}
  \end{aligned}
  \vspace{-.2em}
  \label{eq:ULC-net}
\end{equation}

During the training phase, the overall loss function $\mathcal{L}_{ULC}$ is defined as:
\vspace{-.5em}
\begin{equation}
  \begin{aligned}
    \mathcal{L}_{ULC} = \lambda_{1}\mathcal{L}_{L1} +
    \lambda_{2}\mathcal{L}_{cyc} +
    \lambda_{3}\mathcal{L}_{D}  \text{,}
  \end{aligned}
  \vspace{-.5em}
  \label{eq:ULC-loss}
\end{equation}
where $\lambda_{i}(i=1,2,3)$ are weights of the three loss functions.

\vspace{-1em}
\paragraph{Point-wise L1 Loss.}
The first term $\mathcal{L}_{L1}$ is defined by the point level $l1$ loss function to calculate errors of the landmark coordinates:
\vspace{-.5em}
\begin{equation}
    \mathcal{L}_{L1} = ||\hat{\boldsymbol{l}}_{T,n} - \boldsymbol{l}_{T,n}||_1   \text{.}
    \vspace{-.5em}
    \label{eq:ULC-loss1}
\end{equation}

\vspace{-1em}
\paragraph{Cycle Consistent Loss.}
The second term $\mathcal{L}_{cyc}$ constrains that the converted $\hat{L}_{T,n}$ is capable of converting back again:
\vspace{-.5em}
\begin{equation}
    \mathcal{L}_{cyc} = ||\boldsymbol{\psi}(\boldsymbol{l}_{S,r}, \boldsymbol{\psi}(\boldsymbol{l}_{T,r}, \boldsymbol{l}_{S,n})) - \boldsymbol{l}_{S,n}||_1   \text{,}
    \vspace{-.2em}
    \label{eq:ULC-loss2}
\end{equation}
where $L_{S,r}$ represents the reference expression of person $S$.

\vspace{-1em}
\paragraph{Adversarial Loss.}
Here we regard the ULC $\boldsymbol{\psi}$ as a generator, and the third term $\mathcal{L}_{D}$ contains two discriminators ($D_{TF}$ and $D_{S}$) to make $\boldsymbol{\psi}$ more accurate and robust.
The discriminator $D_{TF}$ is used to judge $real$ or $fake$ of the landmark, and $D_{S}$ is used to estimate the identity similarity score of the landmark pair.
Two discriminator losses are defined as:
\vspace{-.5em}
\begin{equation}
\begin{aligned}
    \mathcal{L}_{D_{TF}} = &\mathbb{E}_{x \sim p_{data}(x)}[\log (D_{TF}(x))] + \\
    &\mathbb{E}_{z \sim p_{data}(z)}[\log (1 - D_{TF}(\boldsymbol{\psi}(z)))],
    \vspace{-.5em}
    \label{eq:ULC-loss3-1}
\end{aligned}
\end{equation}

\vspace{-.5em}
\begin{equation}
\begin{aligned}
    \mathcal{L}_{D_{S}} = &\mathbb{E}_{x_1,x_2 \sim p_{data}(x)}[\log (D_{S}(x_1,x_2))] + \\
    &\mathbb{E}_{z \sim p_{data}(z),x_1 \sim p_{data}(x)}[\log (1 - D_{S}(x_1,\boldsymbol{\psi}(z)))],
    \vspace{-.5em}
    \label{eq:ULC-loss3-2}
\end{aligned}
\end{equation}
where $x$ indicates real landmark data space, and $z$ indicates an input space of $\boldsymbol{\psi}$.

\subsection{Geometry-aware Generator}
Given a target reference face image $\boldsymbol{I}_{T,r}$ and a converted landmark $\hat{\boldsymbol{L}}_{T,n}$, the GAG reenacts the target face $\hat{\boldsymbol{I}}_{T,n}$ which is in the same expression with the source face $\boldsymbol{I}_{S,n}$.
Specifically, the GAG is designed based on the commonly used \emph{Pix2Pix} framework~\cite{pix2pix2017} in a decoupling thought.
It simultaneously learns the appearance information from $\boldsymbol{I}_{T,r}$ and the geometry information from $\hat{\boldsymbol{L}}_{T,n}$ in different paths.
As shown in Figure~\ref{fig:3-FReeNet} (bottom), the GAG consists of an image encoder $\boldsymbol{\phi}_{\theta_1}$, a landmark encoder $\boldsymbol{\phi}_{\theta_2}$, a transformer $\boldsymbol{\phi}_{\theta_3}$, and an image decoder $\boldsymbol{\phi}_{\theta_4}$.
The transformer consists of three ResBlock parts, with each part connecting to the output of the landmark encoder. This kind of design ensures that the geometric information ($\hat{\boldsymbol{L}}_{T,n}$) will be enhanced for the output image, and the process can be described as:

\vspace{-.5em}
\begin{equation}
  \begin{aligned}
    \hat{\boldsymbol{I}}_{T,n} &= \boldsymbol{\phi}(\boldsymbol{I}_{T,r}, \hat{\boldsymbol{L}}_{T,n})\\
    &= \boldsymbol{\phi}_{\theta_4}(\boldsymbol{\phi}_{\theta_3}(\boldsymbol{\phi}_{\theta_1}(\boldsymbol{I}_{T,r}), \boldsymbol{\phi}_{\theta_2}(\hat{\boldsymbol{L}}_{T,n})))  \text{.}
  \end{aligned}
  \vspace{-.2em}
  \label{eq:GAG-net}
\end{equation}

GAG is designed in a decoupling idea and can efficiently reenact identity-reserved and expression-converted face images among multiple persons.
During the training phase of the GAG, the full loss function $\mathcal{L}_{GAG}$ is defined as:
\vspace{-.5em}
\begin{equation}
    \begin{aligned}
        \mathcal{L}_{GAG} = \lambda_{pix}\mathcal{L}_{pix} +
        \lambda_{adv}\mathcal{L}_{adv} +
        \lambda_{TP}\mathcal{L}_{TP} \text{,}
    \end{aligned}
    \vspace{-.5em}
    \label{eq:GAG-loss}
\end{equation}
where $\lambda_{L},\lambda_{adv}$, and $\lambda_{TP}$ represent weight parameters.

\vspace{-1em}
\paragraph{Pixel-wise L1 Loss.} The first term $\mathcal{L}_{pix}$ calculates $l1$ errors between generated and supervised images:
\vspace{-.5em}
\begin{equation}
    \mathcal{L}_{pix} = ||\hat{\boldsymbol{I}}_{T,n} - \boldsymbol{I}_{T,n}||_1 \text{.}
    \vspace{-.5em}
    \label{eq:GAG-loss1}
\end{equation}

\vspace{-1.2em}
\paragraph{Adversarial Loss.} The second term $\mathcal{L}_{adv}$ introduces the discriminator to improve the realism of the generated images in an adversarial idea:
\vspace{-.5em}
\begin{equation}
\begin{aligned}
    \mathcal{L}_{adv} = &\mathbb{E}_{x \sim p_{data}(x)}[\log (D(x))] +  \\
    &\mathbb{E}_{k \sim p_{data}(k), l \sim p_{data}(l)}[\log (1 - D(\boldsymbol{\phi}(k, l)))],
    \vspace{-.5em}
    \label{eq:GAG-loss2}
\end{aligned}
\end{equation}
where $x$ indicates real image data space, and $k$ and $l$ represent input image and landmark space respectively of $\boldsymbol{\psi}$. 
Discriminator $D$ is similar to the work\cite{CycleGAN2017}.

\vspace{-1em}
\paragraph{Triplet Perceptual Loss.} The third term $\mathcal{L}_{TP}$ does duty for intra-class and inter-class evaluations, which helps generate images with more details as well as decouple the appearance and geometry information.
It will be specified in the following section~\ref{sec:TP}.

\begin{figure}[t]
    \centering
    \includegraphics[width=1.0\columnwidth]{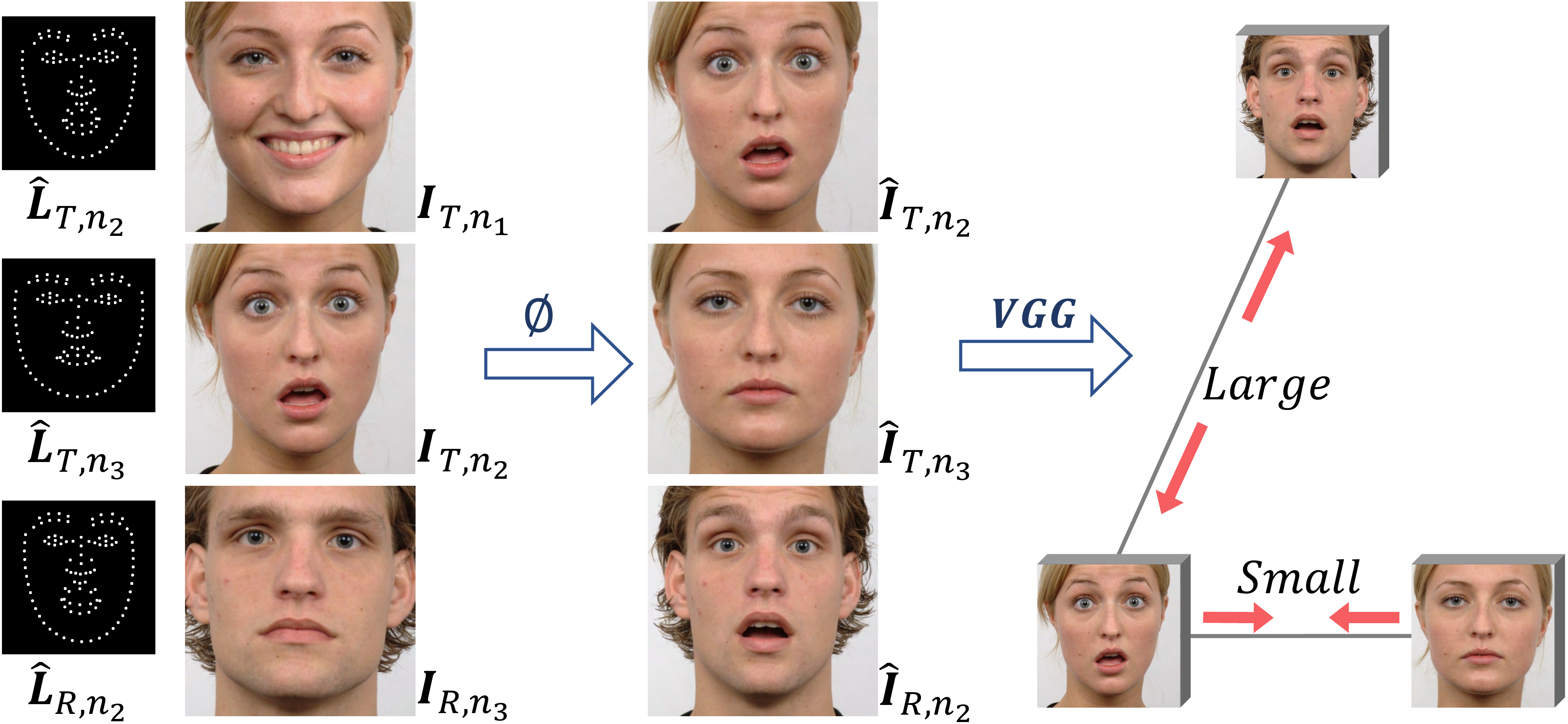}
    \caption{Schematic diagram of the triplet perceptual loss. Simultaneously maximizing inter-class perceptual variation of reenacted images ($\hat{\boldsymbol{I}}_{T,n_2}$ and $\hat{\boldsymbol{I}}_{R,n_2}$) and minimizing intra-class perceptual variation ($\hat{\boldsymbol{I}}_{T,n_2}$ and $\hat{\boldsymbol{I}}_{T,n_3}$).}
    \vspace{-1em}
    \label{fig:3-tripletPerceptionLoss}
\end{figure}

\subsection{Triplet Perceptual Loss}\label{sec:TP}
During the training phase, we find that the GAG module is ill-conditioned to learn a mapping between the input landmark and the generated image if only under the supervision of the adversarial and L1 losses.
This problem is caused by the different distributions between the RGB and landmark images.
The generator tends to only learn from the landmark since its distribution is simple.
In order to conquer this problem, we combine the triplet loss~\cite{schroff2015facenet} and the perceptual loss~\cite{johnson2016perceptual} to form a novel triplet perceptual (TP) loss, which can maximize inter-class and minimize intra-class perceptual variations.

As shown in Figure~\ref{fig:3-tripletPerceptionLoss}, two images $\boldsymbol{I}_{T,n_1}$ and $\boldsymbol{I}_{T,n_2}$ with arbitrary expressions ($n_1$ and $n_2$) are randomly selected within the target person $T$, while the third image $\boldsymbol{I}_{R,n_3}$ is randomly selected within another person $R$ in an arbitrary expression $n_3$.
Images $\hat{\boldsymbol{I}}_{T,n_2},\hat{\boldsymbol{I}}_{T,n_3},\hat{\boldsymbol{I}}_{R,n_2}$ are generated by the GAG with inputs $(\hat{\boldsymbol{L}}_{T,n_2}, \boldsymbol{I}_{T,n_1})$,  $(\hat{\boldsymbol{L}}_{T,n_3}, \boldsymbol{I}_{T,n_2})$, and $(\hat{\boldsymbol{L}}_{R,n_2}, \boldsymbol{I}_{R,n_3})$ respectively.
$\hat{\boldsymbol{L}}_{R,n_2}$ indicates converted landmark image is from identity $T$ to $R$ with expression $n_2$, and so do $\hat{\boldsymbol{I}}_{T,n_2}$ and $\hat{\boldsymbol{I}}_{T,n_3}$.
Then the TP loss is applied to distinguish images generated by similar landmarks but different reference persons denoted as:

\vspace{-1.5em}
\begin{equation}
  \begin{aligned}
    \mathcal{L}_{TP}(\hat{\boldsymbol{I}}_{T,n_2},\hat{\boldsymbol{I}}_{T,n_3},\hat{\boldsymbol{I}}_{R,n_2}) = &\Big[m +
    D\left( \kappa(\hat{\boldsymbol{I}}_{T,n_2}), \kappa(\hat{\boldsymbol{I}}_{T,n_3}) \right) \\
    &- D\left( \kappa(\hat{\boldsymbol{I}}_{T,n_2}), \kappa(\hat{\boldsymbol{I}}_{R,n_2}) \right) \Big]_+ \text{,}
  \end{aligned}
  \vspace{-.2em}
  \label{eq:triplet}
\end{equation}
where $m$ is the margin for controlling intra and inter gaps; $\kappa(\cdot)$ means features extraction operation by VGG~\cite{simonyan2014very}; $D(\cdot,\cdot)$ means L2 distance; $+$ means the value is positive.

Without TP loss, $\hat{\boldsymbol{I}}_{T,n_2}$ is only a converted intra-class image under the supervision of its ground-truth, which means the GAG tends to couple the landmark and the generated face naturally.
By comparison, the GAG has additional inter-class or intra-class constrains for the generated $\hat{\boldsymbol{I}}_{T,n_2}$ from $\hat{\boldsymbol{I}}_{T,n_3}$ and $\hat{\boldsymbol{I}}_{R,n_2}$ when using the TP loss.
In this case, the TP loss forces one landmark to participate in the face reenactment of all target persons in the dataset. Thus the GAG has to extract features from the reference face and the landmark image simultaneously to reenact the target person.

\subsection{Training Scheme}\label{training_scheme}
The training process of the FReeNet consists of two phases.
In the first phase, we train the \emph{ULC} module from scratch with the loss function defined in Eq.~\ref{eq:ULC-loss}, where the corresponding loss weights are set as $\lambda_{1}=100$, $\lambda_{2}=10$, and $\lambda_{3}=0.1$.
Then we fix the parameters of the trained \emph{ULC} module and learn the parameters of \emph{GAG} module in the second phase, where the loss weights $\lambda_{pix}$, $\lambda_{adv}$, and $\lambda_{TP}$ are 100, 1, and 0.1, respectively.
The margin value $m$ of the TP loss is set 0.3 in all experiments.

\section{Experiments}
In this section, we evaluate our approach on the aforementioned RaFD and Multi-PIE datasets and make a contrastive analysis with the state-of-the-art methods.
Moreover, some ablation studies on the RaFD dataset are conducted to illustrate the effect of each proposed component in FReeNet, and additional images in the wild are further tested.
Finally, a series of landmark interpolation and manipulation experiments on the RaFD dataset are performed to highlight the decoupling superiority of our approach.
We also supply a demo video in the supplementary file for more generative results.

\begin{figure*}[ht]
    \centering
    \includegraphics[width=0.98\linewidth]{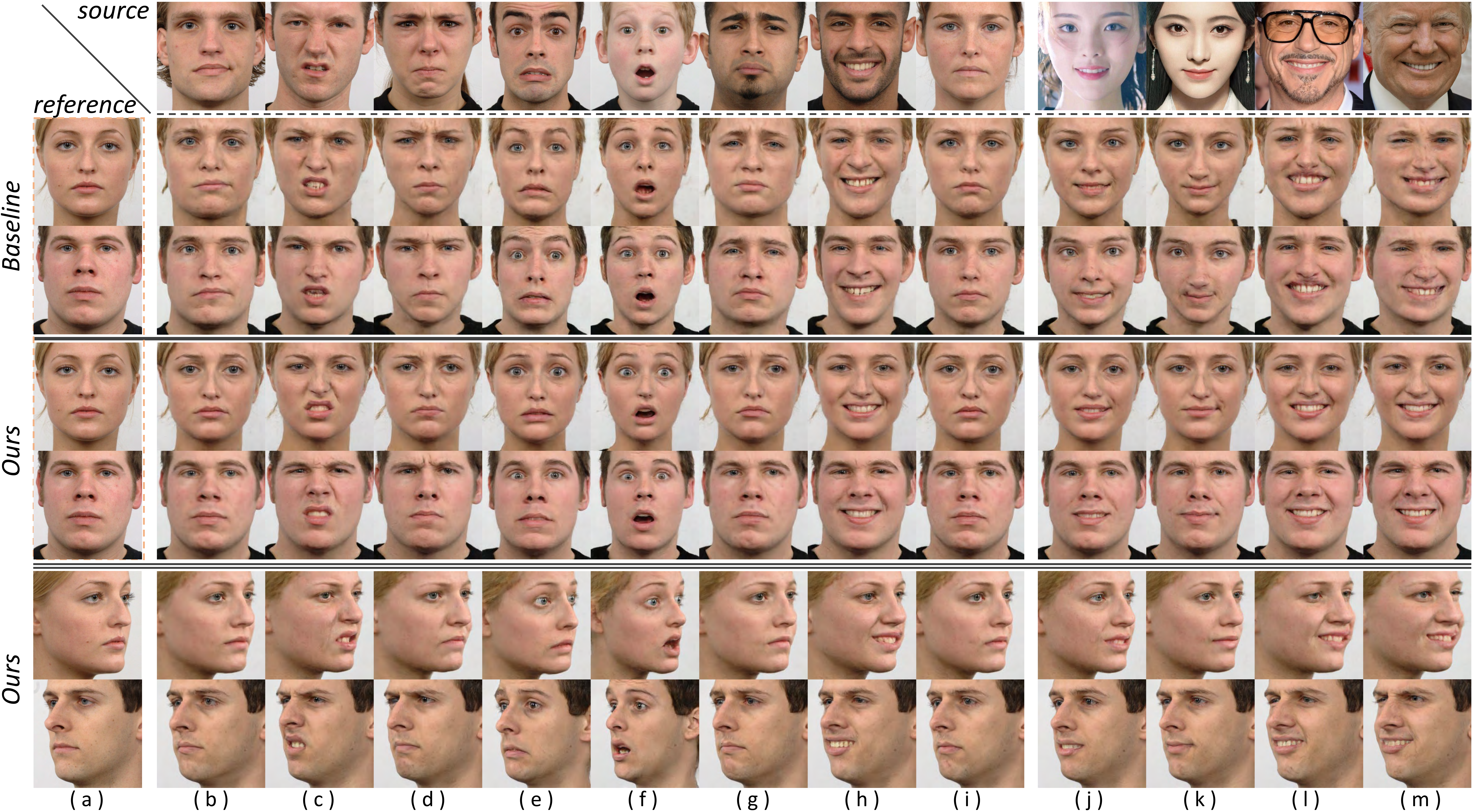}
    \caption{Generative results compared with the baseline on the RaFD dataset. The first column and row are reference and source images, respectively. The right four source images are in the wild. Images on the second and third rows are generated by the baseline, while the other images are generated by our method. Please zoom in for more details.}
    \vspace{-.5em}
    \label{fig:4-compare}
\end{figure*}

\subsection{Datasets and Implementation Details}
\paragraph{RaFD.}
The Radboud Faces Database (RaFD)~\cite{langner2010presentation} consists of 8,040 images collected from 67 participants.
Each participant makes eight facial expressions in three different gaze directions and five different angles, and all $45^{\circ}$, $90^{\circ}$, and $135^{\circ}$ face images are used in the paper.
Images are cropped to $416\times 416$ with face-centered and then resized to $256\times 256$.
The landmark with 106 key points for each face image is provided by HyperLandmark~\cite{guo2019pfld}.
\vspace{-1em}
\paragraph{Multi-PIE.}
A total of 337 subjects (more than 750,000 images) are recorded inside the CMU 3D room using a hardware-synchronized network of 15 high-quality video cameras and 18 flashes.
The detailed processing of faces is similar to the \emph{RaFD} dataset.
\vspace{-1em}
\paragraph{Evaluation Metrics.}
We use \emph{Amazon Mechanical Turk} (AMT) to evaluate the visual quality of reenacted images, \emph{Structural Similarity} (SSIM)~\cite{wang2004image} to measure the structural similarity between generated and real images, and \emph{Fr\'echet Inception Distance} (FID)~\cite{Heusel2017GANsTB} to measure the realism and variation of generated images.

\vspace{-1em}
\paragraph{Implementation Details.}
We follow our training scheme described in Section~\ref{training_scheme}.
For the ULC, we use Adam~\cite{kingma2014adam} optimizer for all modules and set $\beta_1=0.99$, $\beta_2=0.999$.
The initial learning rate is set to $3e^{-4}$, and it decays by ten every 300 epochs.
We train the converter for 1,000 epochs, and the batch size is 16.
For the GAG, we use Adam optimizer and set $\beta_1=0.5$, $\beta_2=0.999$.
The initial learning rate is set to $2e^{-4}$, and it decays by ten every 120 epochs.
We train the converter for 400 epochs, and the batch size is 4.
PatchGAN proposed in~\cite{pix2pix2017} is used as the discriminator, and the training setting is the same as the generator. 
We further test the inference speed of the FReeNet that the ULC can efficiently run with CPU at a speed of 
\emph{878 FPS} and the inference time of the proposed GAG model is around 13.5 ms with a 2080 Ti GPU.
Detailed structure and parameters of FReeNet can be found in the supplementary material.
For the \emph{baseline} in the paper, we choose a modified \emph{Pix2Pix}~\cite{pix2pix2017} that the landmark is seen as the fourth channel concatenated to the input RGB image.

\subsection{Qualitative Results}
We conduct and discuss a series of qualitative experiments on the RaFD and Multi-PIE datasets to demonstrate the high quality of generated images and the flexibility of the proposed framework.
As shown in Figure~\ref{fig:4-compare}, we randomly choose eight identities with different expressions from the training dataset and four identities with random expressions outside the dataset as source images.
Then their facial expressions and movements are transferred to three target persons in three poses. 
The results show that our proposed FReeNet can preserve geometry information of the reference images and reenact high-quality target face images.
For example, the generated images of the baseline at the column (k) are unable to keep the facial contour as the reference images, while our method can achieve it.
Moreover, our model performs better at details such as the upper lip (the column (l)), nose (the column (m)), and teeth (the column (h)).
The generated faces of our method are photorealistic and expression-alike, where the facial appearances and contours are consistent with the reference images.

\begin{figure}[t]
    \centering
    \includegraphics[width=1\linewidth]{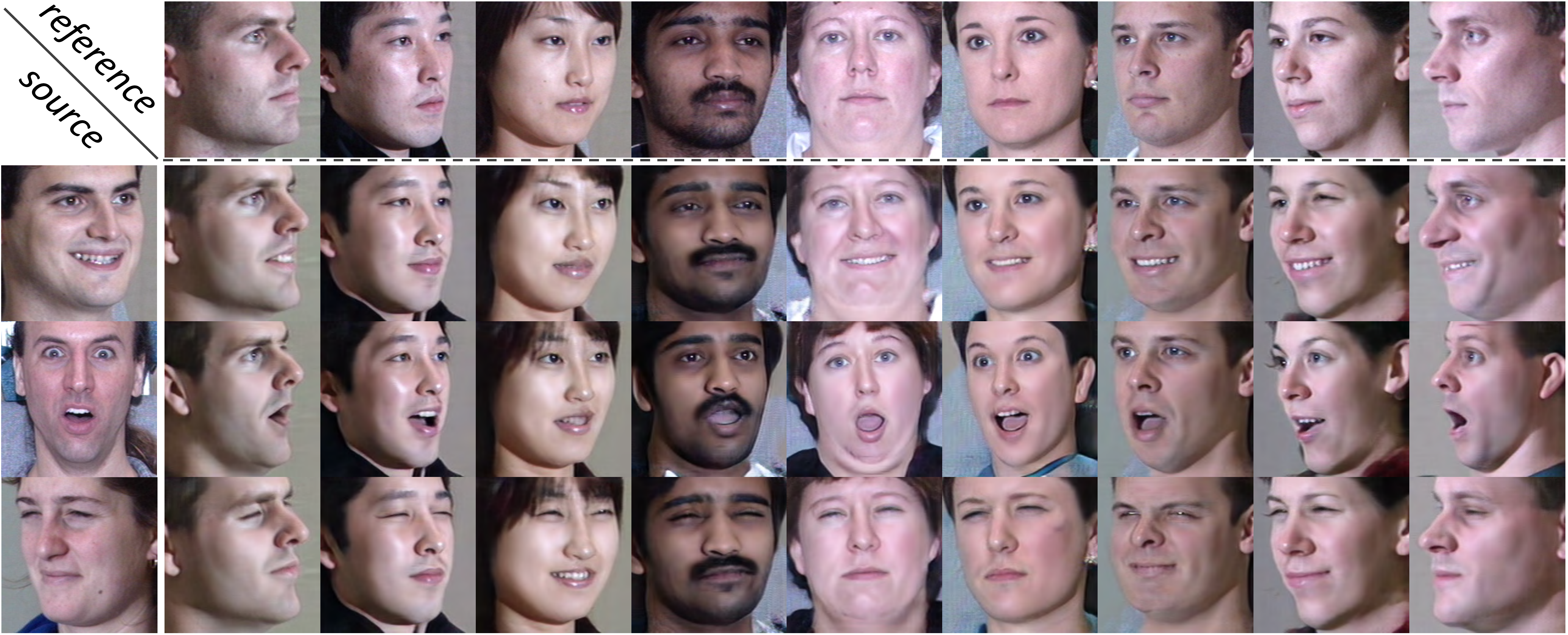}
    \caption{Experimental results in the same illumination on the Multi-PIE dataset. The first column contains three randomly selected source identities with different poses and expressions, and the first row presents nine reference identities in different poses. The rest faces are reenacted by our approach.
    }
    \vspace{-.5em}
    \label{fig:PIE-1}
\end{figure}

\begin{figure}[t]
    \centering
    \includegraphics[width=1\linewidth]{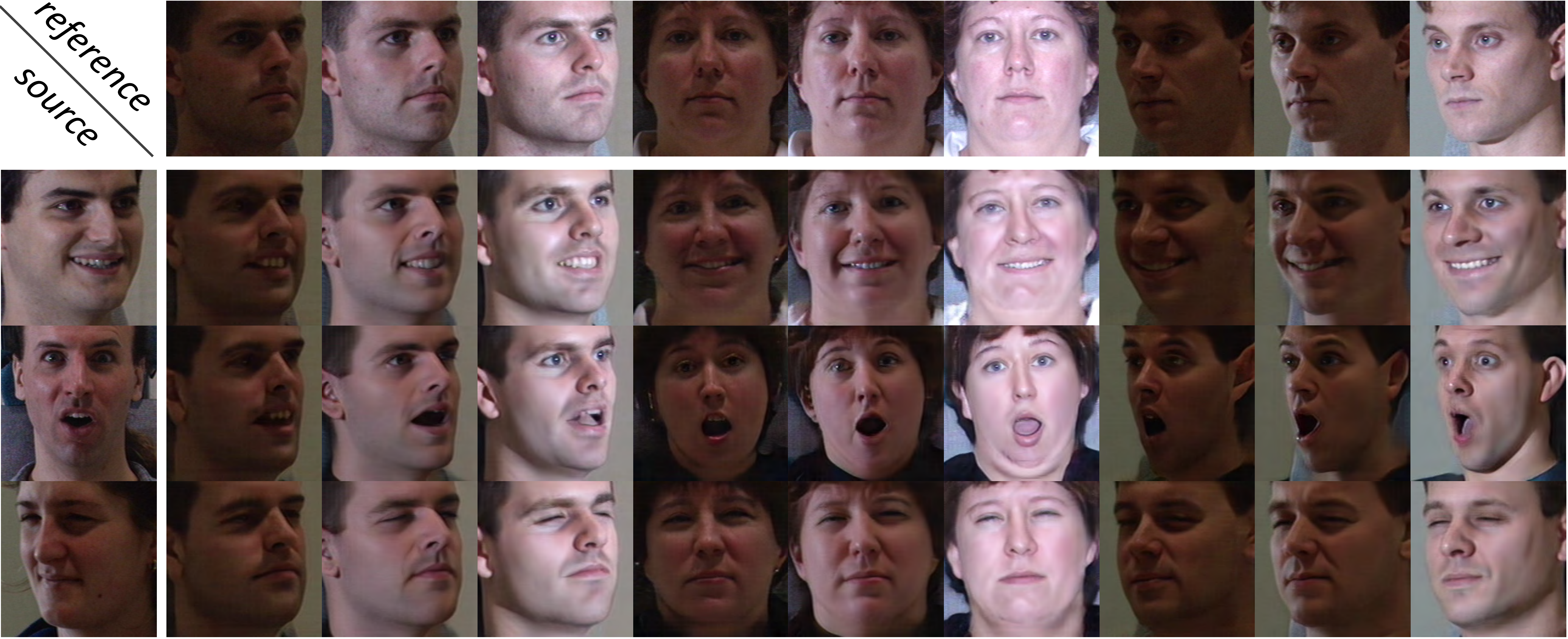}
    \caption{Experimental results in the varying illumination on the Multi-PIE dataset. The first row shows three reference persons in different illumination, and the first column contains three randomly selected source identities with different poses and expressions. The rest faces are reenacted by our approach.
    }
    \vspace{-1em}
    \label{fig:PIE-2}
\end{figure}

Some experiments on the Multi-PIE dataset are further conducted to demonstrate the effectiveness of our proposed method. 
As shown in Figure~\ref{fig:PIE-1} and Figure~\ref{fig:PIE-2}, the experimental results show that our approach can well transfer expression from the source persons to the target persons while simultaneously maintain the same pose and illumination information of the reference images.

\subsection{Quantitative Results}

\begin{table}[t] \small
  \begin{center}
  \caption{Metric evaluation results of the reproduced baseline and our method with different components on the RaFD dataset. Missing entry (-) means that the model is not evaluated by the metric.}
  \label{tab:metrics}
    \begin{tabular}{cccc}
    \toprule
    \noalign{\smallskip}
      \textbf{Model} & \textbf{SSIM} $\uparrow$ & \textbf{FID} $\downarrow$ & \textbf{AMT} \\
    \noalign{\smallskip}
    \midrule
      Pix2Pix~\cite{pix2pix2017}   & 0.629   & 12.84 & 41.3\% \\
      GAG                          & 0.659   & \textbf{11.67} &  - \\
      GAG+ULC                       & \underline{0.711} & 13.26 & - \\
      GAG+ULC+TP (full)   & \textbf{0.717}   & \underline{12.17} & \textbf{74.9\%} \\
    \hline
    \end{tabular}
  \vspace{-10pt}
  \end{center}
\end{table}

\begin{table}[t] \small
  \begin{center}
  \caption{Parameter and speed comparisons of different models when learning all transformations among $n$ persons. Missing entry (-) means that the model has no corresponding component.}
  \label{tab:parameters}
    \begin{tabular}{cccc}
    \toprule
    \noalign{\smallskip}
      \multirow{2}{*}{\textbf{Model}} & \multicolumn{2}{c}{\textbf{Parameters (M)}} & \textbf{Speed} \\
      \cmidrule(l{3mm}r{3mm}){2-3}
      & Transfer & Generator & (FPS) \\
      \hline
      Pix2Pix~\cite{pix2pix2017}   & - & 16.7$\times$n(n-1) & 75 \\
      Xu~\etal~\cite{xu2017face}   & - & 16.7$\times$n(n-1) & 73 \\
      ReenactGAN~\cite{wu2018reenactgan}   & 7.8$\times$n & 61.1$\times$n & 48 \\
      X2Face~\cite{wiles2018x2face}   & - & 108.8 & 16 \\
      Ours                         & 4.5 & 17.3 & 57 \\
      \hline
    \end{tabular}
  \vspace{-20pt}
  \end{center}
\end{table}

We choose SSIM and FID metrics to evaluate the effectiveness of our proposed method on the RaFD dataset quantitatively.
During the experiment, we generate 100 reenacted images for each reference identity (67 identities totally) where corresponding 100 source images are randomly selected from other identities (6,700 images totally). In this way, diversified images can be generated because different identities (used as the source image) have different face attributes, \eg face contour and interorbital distance.
From the comparison results shown in Table~\ref{tab:metrics}, the proposed GAG outperforms the baseline in two metrics.
However, both the two models can not keep the identity consistent for no landmark adaptation operation, which we call an identity shift problem.
So we design the ULC module to alleviate the problem. As a result, the metric scores have a significant improvement in SSIM for the identity preserving capacity of the ULC, but a little descend in FID.
We analyze it reasonable for that the FID metric judges both variety and reality of the image, and GAG model can generate more various images because various contour-inconsistent landmark images of other persons are used for one identity. 
The last row indicates the proposed TP loss brings a little increase in SSIM ($0.006\uparrow$) and an obvious benefit on FID ($0.96\downarrow$) at the same time. 
The reason can be intuitively found from \ref{sec:ablation} that the TP loss boosts the quality of the reenacted faces for having more facial details.

We further perform a user study on Amazon Mechanical Turk (AMT). 
For each of 30 testers, 67 real and 67 fake images of different identities are shown in random order with unlimited decision time. 
The result shows that our generated images confuse testers on 74.9\% trials, \emph{i.e.}, 74.9\% generated images are recognized as real, while this value in the baseline is only 41.3\%. 
As a reference, the percentage of true positive samples is 71.1\%, which is unexpectedly slightly lower than our reenacted images.

Moreover, we compare several most relevant works on model parameters (M: million) and inference speed (FPS: frames per second) to further prove the efficiency of our method, as shown in Table~\ref{tab:parameters}.
The model parameter of our method is much lower than other methods especially when the identity number $n$ is large, because our method only requires one unified model whatever $n$ is that reduces the occupancy of space. Also, our approach has the approximate time consumption (57 FPS) with other methods (\eg Pix2Pix, Xu~\etal, and ReenactGAN) while nearly a third of the time against X2Face (16 FPS) when reenacting the special person by a 2080 Ti GPU, which means that our method is efficient for practical application.

For one identity generation, our approach has the approximate time consumption with other methods(\eg Pix2Pix and ReenactGAN), while takes less than half time against X2Face. For multiple identities generation, our approach uses only one unified model while other methods have to reload the model of the corresponding identity, which consumes extra time and space.

\subsection{Comparison with State-of-the-art}\label{sec:SOTA}
\begin{figure}[t]
    \centering
    \includegraphics[width=1\linewidth]{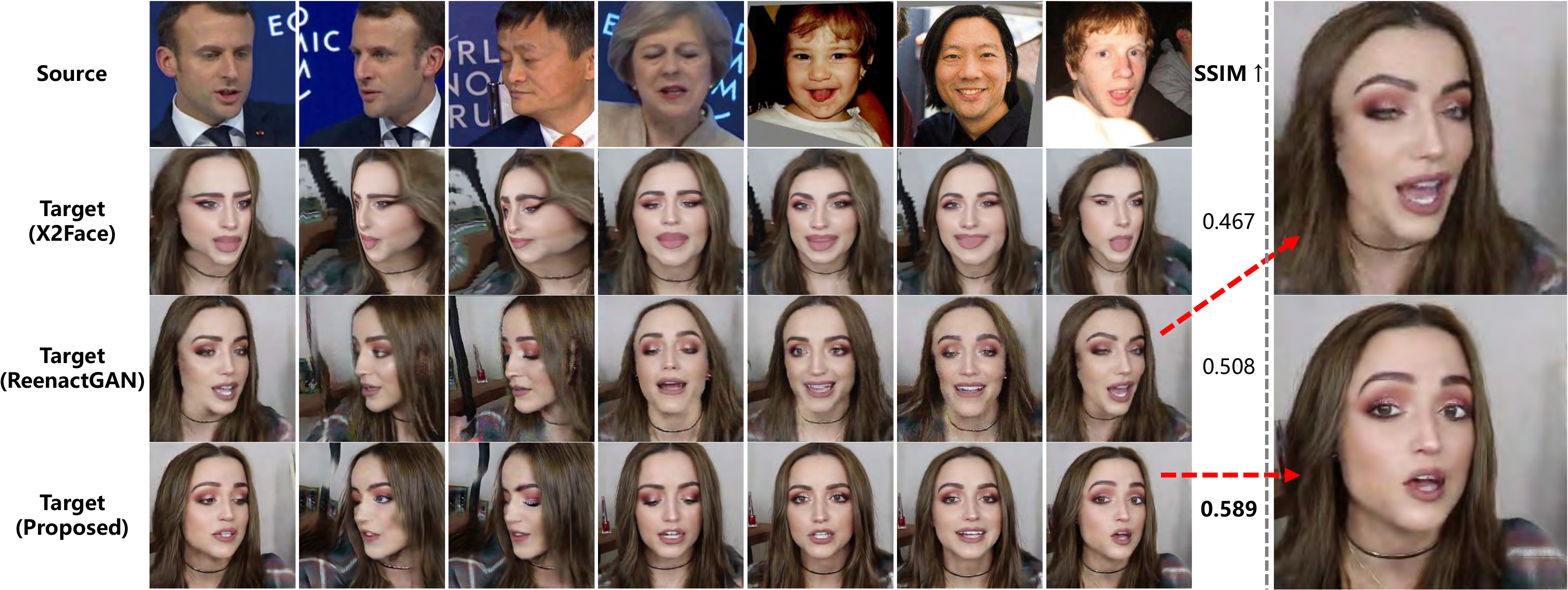}
    \caption{Result comparisons with state-of-the-art methods on the CelebV dataset. The first row is source images. The results of the second and third rows are from X2Face and ReenactGAN. The results of the last row are ours. Please zoom in for more details.}
    \vspace{-1em}
    \label{fig:4-compare_reenactGAN}
\end{figure}

As shown in Figure~\ref{fig:4-compare_reenactGAN}, we also conduct a contrast experiment with most related methods on the CelebV dataset~\cite{wu2018reenactgan} that contains five celebrities.
We first choose 15 images of each person in this dataset to build paired images, which are used to train the ULC module.
The generative results indicate our method can reenact more photorealistic and detail-abundant faces, such as teeth and hair. 
In detail, our approach gains 26.1\% and 15.9\% improvements compared to X2Face and ReenactGAN, respectively.
Note that the \emph{ULC} module is slightly modified to regress $\hat{\boldsymbol{l}}_{T,n}$ that has the same pose with $\boldsymbol{l}_{S,n}$, so as to compare with those methods.

\begin{table}[htp] \small
  \begin{center}
  \caption{ACE results of different loss terms on the RaFD dataset.}
  \label{tab:losses}
    \begin{tabular}{cccc}
    \toprule
      \textbf{Losses} & \textbf{$\mathcal{L}_{L1}$} & \textbf{$+\mathcal{L}_{cyc}$} & \textbf{$+\mathcal{L}_{cyc},\mathcal{L}_{D}$} \\
    \midrule
      ACE & $7.236\pm0.015$ & $4.526\pm0.015$ & $0.895\pm0.010$ \\
    \hline
    \end{tabular}
    \vspace{-10pt}
    \end{center}
\end{table}

\subsection{Ablation Study}\label{sec:ablation}

\paragraph{Loss Functions of the ULC.}
We test the average coordinate-wise error (ACE) of the converted landmarks in different loss functions when training the ULC module. 
As shown in Table~\ref{tab:losses}, ACE value gradually decreases when different loss terms are added, which proves the effectiveness of discriminators.

\vspace{-1em}
\paragraph{Effect of the TP Loss.}

\begin{figure}[t]
    \centering
    \includegraphics[width=1.0\columnwidth]{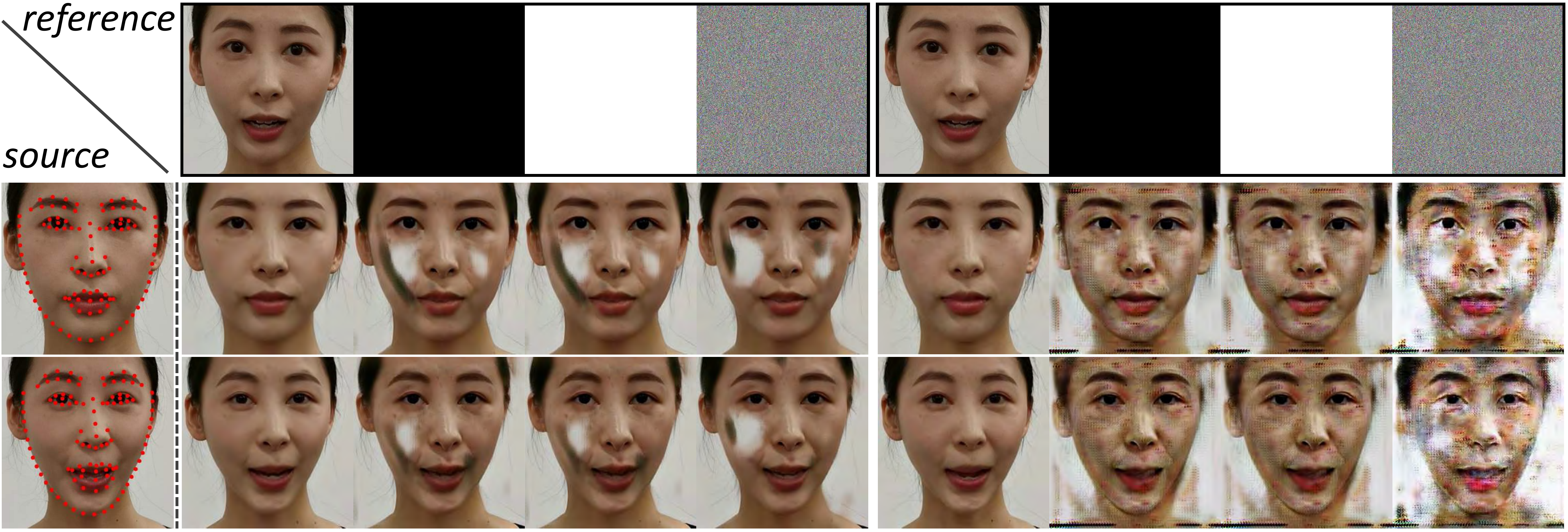}
    \caption{A toy experiment for testing the effect of \emph{TP loss}. The second and third rows are in different source images. The second to fifth columns are results without TP loss while the rest columns use TP loss. Please zoom in for more details.}
    \vspace{-.2em}
    \label{fig:4-seperationExperiment}
\end{figure}

To further illustrate the effectiveness of the \emph{TP} loss, we conduct experiments that only contain two persons, because the triplet loss requires at least two persons. 
As shown in Figure~\ref{fig:4-seperationExperiment}, experimental results show that the \emph{TP} loss can well separate appearance and geometry information to a certain extent.
For example, when feeding the reference image with a black or gaussian noise image, the reenacted face with TP loss contains more abstract features rather than nearly the full face.
It means the GAG itself contains less appearance information and can capture more appearance features from the reference image.

\vspace{-1em}
\paragraph{Components of the FReeNet.}

\begin{figure}[t]
    \centering
    \includegraphics[width=1\linewidth]{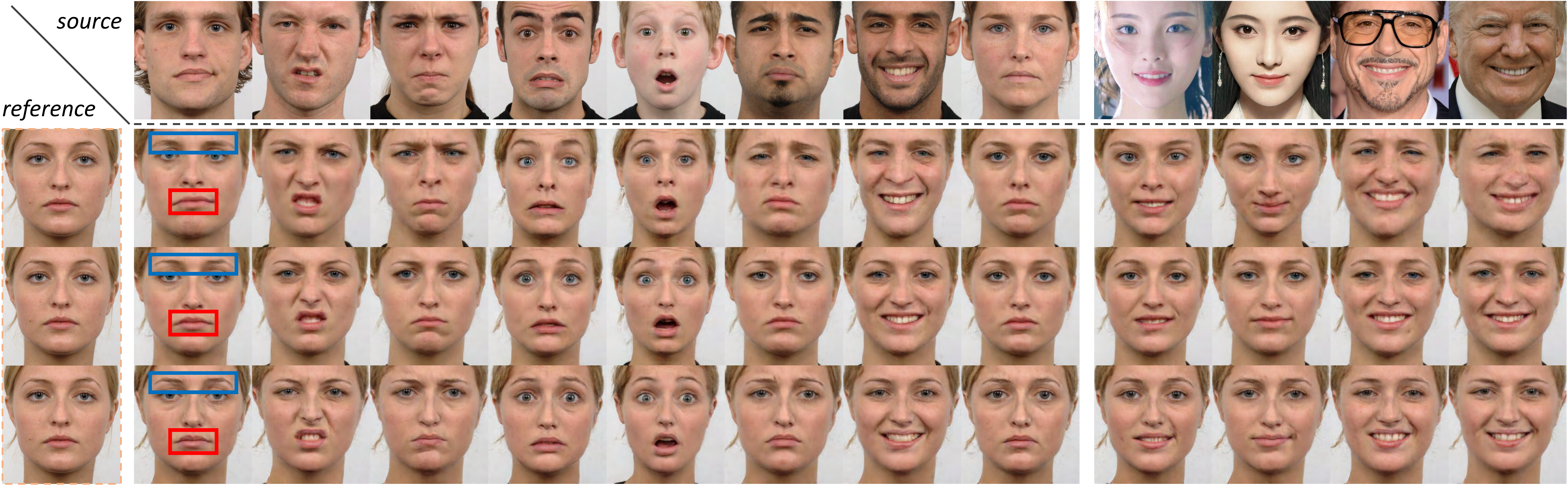}
    \caption{Ablation study on the RaFD dataset. The results of the second row are only generated by GAG. The results after adding ULC (GAG+ULC) are shown in the third row, and the last row shows the results by our complete method (GAG+ULC+TP).
    Please zoom in the blue and red rectangles for more details.
    }
    \vspace{-.2em}
    \label{fig:4-ablation}
\end{figure}

We conduct an ablation study to evaluate the impact of each component on our proposed approach.
As shown in Figure~\ref{fig:4-ablation}, evaluation results of models with different components are reported. 
The first row shows that the proposed GAG can generate images in a good-quality whether the source images are in the dataset or not, but it is unable to preserve the geometry information of the target person. 
Comparing the results of the second and the third rows, we can observe that adding ULC module can significantly enhance the performance. 
The SSIM score meanwhile increases by a large margin, as shown in Table~\ref{tab:metrics}.
Moreover, the effectiveness of the proposed TP loss is evaluated in the last row. 
It shows that the generated images can maintain more opulent facial details, e.g., brow, wrinkle, and mouth, and also have a better score on FID.

\begin{figure}[t]
    \centering
    \includegraphics[width=1\linewidth]{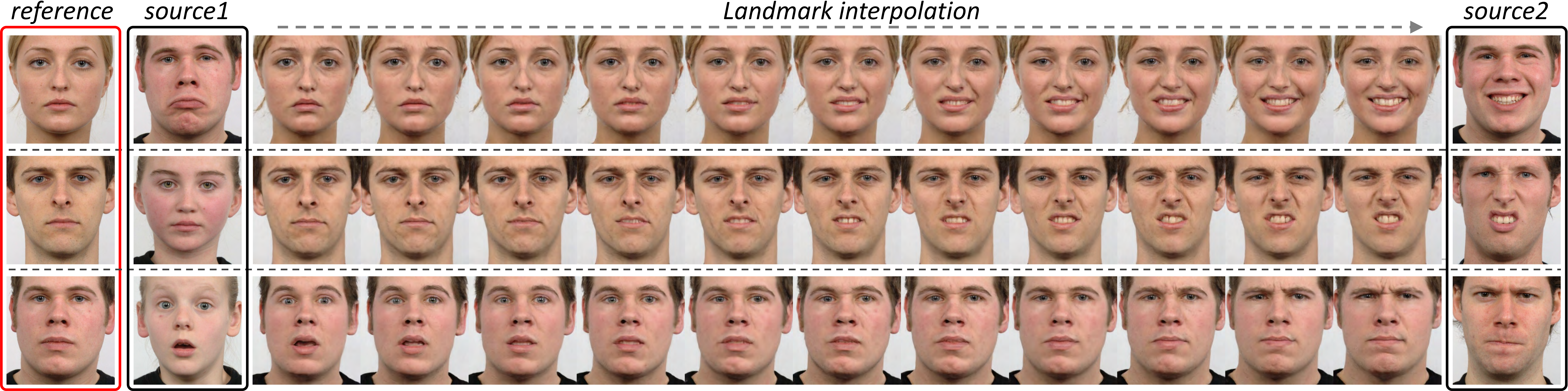}
    \caption{Landmark interpolation experiment on the RaFD dataset. The first column contains three different reference images. The second and last column images of each row are two source images. The rest images are generated by feeding the interpolated landmark image and the corresponding reference image.
    Please zoom in for more details.}
    \vspace{-.3em}
    \label{fig:4-interpolation}
\end{figure}

\begin{figure}[t]
    \centering
    \includegraphics[width=1\linewidth]{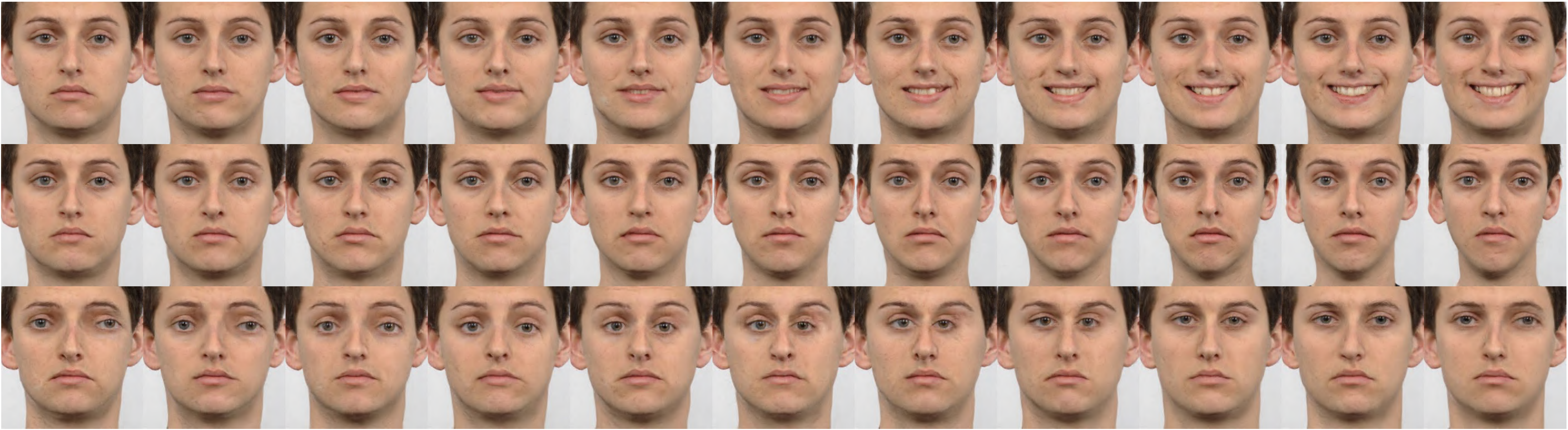}
    \caption{Landmark manipulation experiment on the RaFD dataset. From top to bottom, we manipulate mouth shape (close to open), facial contour (wide to thin), and eye position (rotation) when reenacting target faces.
    Please zoom in for more details.}
    \vspace{-1em}
    \label{fig:4-manipulation}
\end{figure}

\subsection{Landmark Interpolation and Manipulation}
We also present a series of additional qualitative experiments to highlight the advantage of the decoupling design of our model in the RaFD dataset. 
Specifically, we bring two convenient ways to control generated faces: landmark interpolation and landmark manipulation. 
In contrast to traditional interpolation operating in pixel space, the proposed approach interpolates images in latent landmark space.
Specifically, we extract two original landmarks from two source images and get a series of landmarks through interpolating two landmarks. 
Then we can reenact a series of interpolated target images by interpolated landmarks, as shown in Figure~\ref{fig:4-interpolation}.

The other way is directly manipulating point coordinates of the input landmark, which provides a flexible way to adjust the geometric position for the reenacted image.
As shown in Figure~\ref{fig:4-manipulation}, three manipulation experiments are performed to change partial attributes of the generated faces, such as mouth shape, facial contour, and eye position. 
The results show that our approach can keep the identity and other attributes of the reference person when manipulating a specific attribute, which intuitively confirms the advantage of the decoupling idea.

\section{Conclusion}
In this paper, we propose a novel FReeNet to address the multi-identity face reenactment task, which aims at transferring facial expressions from source persons to target persons while keeping the identity and pose consistency to the reference images.
Specifically, a ULC module is proposed to effectively convert the expression of an arbitrary source person to the target person in the latent landmark space.
Then the GAG module input the reference image and the converted landmark image to reenact photorealistic target image.
Moreover, a TP loss is proposed to help the GAG to decouple geometry and appearance information as well as generate detail-abundant faces.
Extensive experiments demonstrate the efficiency and flexibility of our approach. 

We hope our work will help users to achieve more effective and efficient works in the face reenactment task. And our approach can be easily transferred to other domains, such as gesture migration or posture migration of the body.

\myparagraph{Acknowledgment} We thank anonymous reviewers for their constructive comments. This work is partially supported by the National Natural Science Foundation of China (NSFC) under Grant No. 61836015 and Key R\&D Program Project of Zhejiang Province (2019C01004).

{\small
\bibliographystyle{ieee_fullname}
\bibliography{egbib}
}

\end{document}